\documentclass[12pt]{article}
\usepackage{amssymb,amsmath,cite,bm}
\usepackage[dvips]{graphicx,color}
\usepackage{psfrag,subfigure}
\newtheorem{remark}{Remark}
\newtheorem{proposition}{Proposition}

\newcommand{\diff}{\frac{\rm d}{{\rm d}t}}

\newcommand{\norm}[1]{\left\Vert#1\right\Vert}

\begin{document}

% ----------------------------------------------------------------
\title{\bf Lidar SLAM for Autonomous Driving Vehicles}

\author{Farhad Aghili\thanks{email: faghili@encs.concordia.ca}}

\date{}

\maketitle

\begin{abstract}
This paper presents Lidar-based Simultaneous Localization and Mapping (SLAM) for autonomous driving vehicles. Fusing data from landmark sensors and a strap-down Inertial Measurement Unit (IMU) in an adaptive Kalman filter (KF) plus the observability of the system are investigated. In
addition to the vehicle's states and landmark positions, a self-tuning filter estimates the IMU calibration parameters as well as the covariance of the measurement noise. The discrete-time covariance matrix of the process noise, the state transition matrix, and the observation sensitivity matrix are derived in closed-form making them suitable for real-time implementation. Examining the observability of the 3D SLAM system leads to the conclusion that the system remains observable upon a geometrical condition on the alignment of the landmarks.
\end{abstract}

%------------------------------------------------------
\section{Introduction}
%------------------------------------------------------

To measure the pose of a vehicle with high bandwidth and long-term
accuracy and stability usually involves data fusion of different
sensors because there is no single sensor to satisfy both
requirements \cite{Aghili-2010s}. Inertial navigation systems where rate gyros and
accelerations are integrated provide a high bandwidth pose
measurement. However, long-term stability cannot be maintained
because the integration inevitably results in quick accumulation of
the position and attitude errors. Therefore, inertial systems
require additional information about absolute position and
orientation to overcome long-term
drift~\cite{Barshan-Durrant-Whyte-1995,Aghili-Salerno-2016}.

Using IMU and wheel encoders to obtain close estimate of robot
position has been proposed
\cite{Vaganay-Aldon-Fourinier-1993,Aghili-Salerno-2011,Dissanayake-Sukkarieh-Nebot-Durrant-Whyte-2001,Aghili-Salerno-2010}.
The application of these techniques for localization of outdoor
robots is limited, particularly when the robot has to traverse an
uneven terrain or loose soils. Integrating data from a differential
Global Positioning System (GPS) and IMU in an elaborate Kalman
filter can bound the error build-up at low frequency and prevent
drift~\cite{Choi-Park-Kim-2005,Aghili-Salerno-2009}. This method,
however, is not applicable for localization of a rover in a
GPS-denied environment or for the planetary exploration. Other
research focuses on using vision system as the absolute sensing
mechanism required to update the prediction position obtained by
inertial measurements
\cite{Strelow-Singh-2003,Aghili-Parsa-2009,Mallet-Lacroix-Gallo-2000,Aghili-Kuryllo-Okouneva-English-2010a,Lamon-Siegwart-2007,Aghili-2016c,Mourikis-Trawny-Roumeliotis-Helmick-Matthies-2007,Aghili-2010f}.
Vision system and IMU are considered complementary positioning
systems. Although vision systems provide low update rate, they are
with the advantage of long-term position accuracy \cite{Aghili-Kuryllo-Okouneva-English-2010c,Aghili-Parsa-Martin-2008a,Aghili-Kuryllo-Okuneva-McTavish-2009,Aghili-Parsa-2007b}. Hence, fusion of
vision and inertial navigation data, which are, respectively,
accurate at low and high frequencies makes sense. Additionally,
integration of the inertial data continuously provides pose
estimation even when no landmark is observable or the vision system
is temporally obscured. Most vision-based navigation systems work
based on detecting several landmarks along which the vehicle pose is
estimated. The challenge for localization of a vehicle traversing an
unstructured environment is that the map of landmarks is not a
priori known.

The SLAM is referred to the capability to construct a map
progressively in unknown environment being traversed by a vehicle
and, at the same time, to estimate the vehicle pose using the map.
In the past two decades, there have been great advancements in
solving the SLAM problem together with compelling implementation of
SLAM methods for field robotics
\cite{Smith-Self-Cheesman-1987,Durant-Whyte-1988,Ayache-Faugeras-1989,Leonard-Durrant-Whyte-1991,Bailey-Durrant-Whyte-2006}.
Among other methods, the extended KF
 based SLAM has gained
widespread acceptance in the robotic
community~\cite{Leonard-Durrant-Whyte-1991,Xi-Guo-Sun-Huang-2008}.
Nevertheless, nonlinear observers for position and attitude
estimation are also proposed
\cite{Rehbinder-Ghosh-2003,Thienel-Sanner-2003,Vasconcelos-Cunha-Silvestre-2007},
and observability conditions based on the original nonlinear system
are studied
\cite{Rehbinder-Ghosh-2003,Lee-Wijesoma-Guzman-2006,Perera-Melkumyan-2009}.
In particular, it was shown in
\cite{Vasconcelos-Cunha-Silvestre-2007} that based on the landmark
measurements and velocity readings, the convergence of the nonlinear
observer is guaranteed for any initial condition outside a set of
zero measurement. These nonlinear observers, however, assume a
deterministic system, whereas the actual system is stochastic.
Although the effects of bounded disturbances and noises on the
observers for nonlinear systems have been investigated
\cite{Marino-Santosuesso-Tomei-1999,Zhang-2002}, only Kalman filters
are able to optimally reduce the noise in the estimation process.

Despite SLAM methods have reached a level of significant maturity,
their mathematical frameworks are predominantly developed for
two-dimensional planar environments. Observability analysis of the
for 2-dimensional SLAM problem has been studied in the literature
\cite{Chen-Jiang-Hung-1990,Andrade-Cetto-Sanfeliu-2005,Lee-Wijesoma-Guzman-2006,Huang-Dissanayake-2006,Huang-Mourikis-Roumeliotis-2008,Perera-Melkumyan-2009}.
3D SLAM and its implementation for mobile robots and airborne
applications have been been proposed
\cite{Surmann-Nuchter-Lingemann-Hetzberg-2004,Kim-Sukkarieh-2004,Weingarten-Siegwart-2005,Cole-Newman-2006,Nuechter-Lingemann-2006,Kim-Sukkarieh-2007,Abdallah-Asmar-Zelek-2007,Bryson-Sukkarieh-2008,Liu-Hu-Uchimura-2009,Nemra-Aouf-2009}.
The observability of 3D SLAM has been also investigated in
\cite{Bryson-Sukkarieh-2008,Nemra-Aouf-2009}. However, 3D SLAM by
integrating landmark sensors and IMU and the observability analysis
of such a system is not addressed in these references.

This work is aimed at investigating at 3D Simultaneous Localization
and Mapping and its corresponding observability analysis by fusing
data from a 3D Camera and strap-down IMU for autonomous driving vehicles \cite{Aghili-2010s}. Since no
wheel odometry is used in this methodology, it is applicable to
terrestrial and aerial vehicles alike. The IMU calibration
parameters in addition to the covariance matrix of the noise
associated with the measurement landmarks' relative positions  are
estimated so that the KF filter is continually ``tuned'' as well as
possible. The observability of such a technique for 3D SLAM is
investigated and the observability condition base on the number of
fixed landmarks is derived.

\section{Mathematical Model}
%------------------------------------------------------
\subsection{Observation} \label{sec:modeling}
%------------------------------------------------------

\begin{figure}[t]
\centering
\includegraphics[width=10cm]{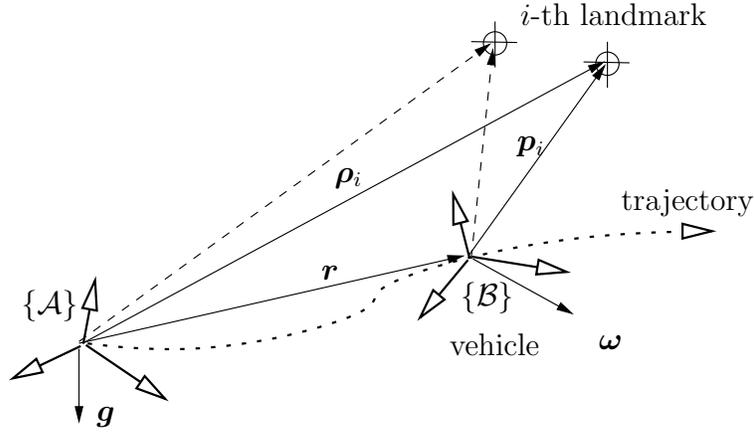} \caption{Coordinate frames for 3D localization of a vehicle.}
\label{fig:frames}
\end{figure}

Fig.\ref{fig:frames} illustrates the coordinate frames which are
used to express the locations of a vehicle and landmarks. Coordinate
frame $\{ {\cal B} \}$ is attached to the vehicle, while $\{ {\cal
A} \}$ is the inertial frame. We assume that coordinate frames $\{
{\cal B} \}$ and  $\{ {\cal A} \}$ are coincident at $t=0$.
Moreover, without loss of generality, we assume that $\{ {\cal B}
\}$ represents the frame of resolution of a 3-D camera system as
well as the IMU coordinate frame. The attitude of a rigid body
relative to the specified inertial frame can be represented by the
unit quaternion $\bm q^T=[\bm q_v^T \; q_o]$, where subscripts $_v$
and $_o$ denote the vector and scalar parts of the quaternion.

By definition, $\bm q_v = \bm e \sin \frac{\vartheta}{2}$ and $q_o =
\cos \frac{\vartheta}{2}$, where $\bm e$ is a unit vector, known as
the Euler axis, and $\vartheta$ is  a rotation angle about this
axis. Below, we review some basic definitions and properties of
quaternions used in the rest of the paper.  Consider quaternions
$\bm q_1$, $\bm q_2$, and $\bm q_3$ and their corresponding rotation
matrices $\bm A_1$, $\bm A_2$, and $\bm A_3$. Then,
\begin{equation*}
\bm A_3= \bm A_1 \bm A_2\quad\Longleftrightarrow\quad \bm q_3 = \bm
q_2 \otimes \bm q_1
\end{equation*}
where $\bm q_3$ is obtained from the quaternion product. The
quaternion product $[\bm q\otimes]$ is defined, analogous to the
cross-product matrix, as
\begin{equation*}
[\bm q\otimes] = \begin{bmatrix} q_o \bm 1_3 - [\bm q_v \times] & \bm q_v \\
-\bm q_v^T & q_o
\end{bmatrix}.
\end{equation*}
Also, the conjugate\footnote{$q_o^*=q_o$ and $\bm q_v^*=-\bm q_v$.}
$\bm q^*$ of a quaternion is defined such that $\bm q^* \otimes \bm
q = \bm q\otimes \bm q^*=[0\;0\;0\;1]^T$. Let us assume that there
exists a set of $m>2$ landmarks which have a fixed position in the
inertial frame, i.e.,
\begin{equation} \label{eq:dot_rho}
\dot {\bm\rho}_i = \bm 0_{3\times 1} \qquad \forall i=1,\cdots ,m
\end{equation}
The position of the landmarks in the vehicle frames, $\{{\cal B}
\}$, are denoted by vectors $\bm p_1 , \cdots \bm p_m$. The
position and orientation of the vehicle with respect to the
inertial frame are represented by  vector $\bm r$ and unit
quaternion $\bm q$, respectively. Apparently, from
Fig.~\ref{fig:frames}, we have
\begin{equation} \label{eq:pi}
\bm p_i =  \bm A^T(\bm q) \big(\bm\rho_i - \bm r \big), \qquad
\forall i=1, \cdots, m.
\end{equation}
where $\bm A(\bm q)$ is the rotation matrix of frame $\{{\cal B}\}$
with respect to frame $\{{\cal A}\}$ that is related to the
corresponding quaternion by
\begin{equation} \label{eq:R}
\bm A(\bm q) = (2q_o^2-1) \bm 1_3 + 2q_o [\bm q_v \times] + 2 \bm
q_v \bm q_v^T.
\end{equation}

%=============================================================
%=============================================================

As suggested in \cite{Andrade-Cetto-Sanfeliu-2005}, we assume that
the map is anchored to a set of $r$ landmarks observed at time
$t=0$. Without loss of generality, we can say that the initial pose
is given by $\bm r(0)=\bm 0_{3 \times 1}$ and $\bm A(0) = \bm 1_3$.
Thus
\begin{equation} \notag
\bm\rho_j = \bm p_j(0) \qquad \forall j=1,\cdots,r.
\end{equation}
The measurement vector includes the outputs of the landmark sensor,
i.e.,
\begin{equation} \label{eq:z}
\bm z =  \begin{bmatrix} \bm p_1 \\ \vdots \\ \bm p_m
\end{bmatrix} + \bm v,
\end{equation}
where vector $\bm v$ is landmark sensor noise which is assumed to be
with covariance matrix
\begin{equation} \label{eq:Rk}
\bm R = E[\bm v \bm v^T ].
\end{equation}
As will be later discussed in Section~\ref{sec:noise-adaptive}, the
Kalman filter tries to estimate the covariance matrix.

Since out of the set of $m$ landmarks there are $r$ known landmarks,
there are $n=m-r$ landmarks with unknown position. Therefore, the
entire state vector to be estimated is:
\begin{equation} \label{eq:states}
\bm x = [\bm q_v^T  \; \bm r^T \; \dot{\bm r}^T \;\bm b_g^T \; \bm
b_a^T \; \bm\rho_{r+1}^T \; \cdots \; \bm\rho_m^T ]^T \in
\mathbb{R}^{15+3n},
\end{equation}
where $\bm b_g$ and $\bm b_a$ are the gyro and accelerometer biases
as will be later discussed in the next section. In view of kinematic
relations \eqref{eq:pi}, we can rewrite \eqref{eq:z} as a function
of the states by
\begin{subequations} \label{eq:obs_nonlin}
\begin{equation} \label{eq:zk}
\bm z = \bm h(\bm x) + \bm v,
\end{equation}
where
\begin{equation} \label{eq:h}
\bm h(\bm x) = \begin{bmatrix} {\bm A}^T(\bm q) \big(\bm\rho_1 - \bm r \big) \\ \vdots \\
{\bm A}^T(\bm q) \big(\bm\rho_n - \bm r \big)
\end{bmatrix}.
\end{equation}
\end{subequations}
constitutes the nonlinear observation model. To linearize the
observation vector, we need to derive the sensitivity of the
nonlinear observation vector with respect to the system variables.
Consider small orientation perturbations
\begin{equation}\label{eq:delta_q1} 
\delta \bm q = \bm q \otimes \bar{\bm q}^*.
\end{equation}
around a nominal quaternion $\bar{\bm q}$---in the following, the
bar sign stands for nominal value. Now, by virtue of $\bm A(\bm q)
=\bm A(\delta \bm q \otimes \bar{\bm q})$, one can compute the
observation vector \eqref{eq:h} in terms of the perturbation $\delta
\bm q$. Using the first order approximation of nonlinear matrix
function $\bm A^T(\delta \bm q)$ from expression \eqref{eq:R} by
assuming a small rotation $\delta \bm q$, i.e., $\|\delta \bm q_v\|
\ll 1$ and $\delta q_0\approx 1$, we have
\begin{equation} \label{eq:A_delq}
\bm A (\delta \bm q) \approx \bm 1_3 + 2[\delta \bm q_v \times ].
\end{equation}
Therefore, by the first-order approximation, the observation vector
can be written as as the following bilinear function
\begin{equation} \notag
\bm h =
\begin{bmatrix}\big( \bm 1_3 - 2[\delta \bm q_v \times] \big) \bar{\bm A}^T \big( \bm\rho_1 - \bm r \big) \\
\vdots \\ \big( \bm 1_3 - 2[\delta \bm q_v \times] \big) \bar{\bm
A}^T \big( \bm\rho_n - \bm r \big)
\end{bmatrix} + \text{HOT}.
\end{equation}
Thus, the observation sensitivity matrix $\bm H= \left.
\frac{\partial \bm h}{\partial \bm x} \right|_{\bm A =\bar{\bm A}}
\in \mathbb{R}^{3m \times(15+3n)}$ can be written as
\begin{equation} \label{eq:H}
\bm H  =
\begin{bmatrix}
2 [\hat{\bm p}_1 \times] & - \bar{\bm A}^T & \bm 0_3 & \bm 0_3 & \bm 0_3 & \bm 0_3 & \cdots & \bm 0_3 \\
\vdots & \vdots & \vdots  &
\vdots & \vdots & \vdots & \vdots & \vdots \\
2 [\hat{\bm p}_r \times] & - \bar{\bm A}^T & \bm 0_3 & \bm 0_3 & \bm 0_3 & \bm 0_3 & \cdots & \bm 0_3 \\
2 [\hat{\bm p}_{r+1} \times] & - \bar{\bm A}^T & \bm 0_3 & \bm 0_3 & \bm 0_3 & \bar{\bm A}^T & \cdots & \bm 0_3 \\
\vdots & \vdots & \vdots  &
\vdots & \vdots & \vdots & \ddots & \vdots \\
2 [ \hat{\bm p}_m \times] & - \bar{\bm A}^T &  \bm 0_3 & \bm 0_3 &
\bm 0_3 & \bm 0_3 & \cdots & \bar{\bm A}^T
\end{bmatrix}
\end{equation}
where
\begin{equation}
\hat{\bm p}_i = \bar{\bm A}(\hat{\bm\rho}_i - \hat{\bm r}) \qquad
\forall i=1,\cdots m
\end{equation}
are the estimation of the landmark positions calculated from the
state estimation.

%------------------------------------------------------
\subsection{Motion Dynamics} \label{sec:dynamics}
%------------------------------------------------------
Denoting the angular velocity of the vehicle by $\bm\omega$, the
relation between the time derivative of the quaternion and the
angular velocity can be readily expressed by
\begin{equation} \label{eq:dot_q}
\dot{\bm q} = \frac{1}{2} \underline{\bm\omega}  \otimes \bm q \quad
\mbox{where} \quad \underline{\bm\omega}=\begin{bmatrix} \bm\omega \\
0
\end{bmatrix}
\end{equation}
The angular rate is obtained from the rate gyro measurement
\begin{equation}
\bm\omega  = \bm\omega_g + \bm b_g  + \bm\epsilon_g
\end{equation}
where $\bm b_g$  is the corresponding bias vector and
$\bm\epsilon_g$ is the angular random walk noises with covariances
$E[\bm\epsilon_g \bm\epsilon_g^T]=\sigma_g^2 \bm 1_3$. The gyro bias
is modeled as
\begin{equation} \label{eq:dotb_g}
\dot{\bm b}_g = \bm\epsilon_{b_{g}},
\end{equation}
where $\bm\epsilon_{b_{g}}$ is the random walk with covariances
$E[\bm\epsilon_{b_{g}} \bm\epsilon_{b_{g}}^T]=\sigma_{b_{g}}^2 \bm
1_3$.

A measurement of the linear acceleration of the vehicle can be
provided by an accelerometer. However, accelerometers cannot
distinguish between the acceleration of gravity and inertial
acceleration. Therefore, we must compensate the accelerometer output
for the effects of gravity. Moreover, the accelerometer output $\bm
a\in\mathbb{R}^3$ is accompanied by its own bias $\bm b_a$, which is
modeled as $\dot{\bm b}_a = \bm\epsilon_{b_a}$ and
$E[\bm\epsilon_{b_a} \bm\epsilon_{b_a}^T] =\sigma_{b_a} \bm 1_3$.
Thus, in the presence of a gravity field, the accelerometer output
equation is represented by
\begin{equation}\label{eq:diff_v}
\ddot{\bm r} = \bm A(\bm q) \big( \bm a +  \bm b_a + \bm\epsilon_a
\big) - \bm g ,
\end{equation}
where  $\bm a$ is acceleration output, accelerometer noise
$\bm\epsilon_a$ is the assumed to be random walk noise
$\bm\epsilon_a$ with covariance $E[\bm\epsilon_a
\bm\epsilon_a^T]=\sigma_a^2 \bm 1_3$, and $\bm g$ is the consrtant
gravity vector expressed in the inertial frame $\{ {\cal A} \}$.

Although the states can be propagated by solving the nonlinear
dynamics equations \eqref{eq:dot_rho}, \eqref{eq:dot_q},
\eqref{eq:dotb_g} and \eqref{eq:diff_v}, the state transition matrix
of the linearized dynamics equations will be also needed to be used
for covariance propagation of the KF. In order to linearize the
equations, consider small position and orientation perturbations
\begin{subequations}
\begin{align} \label{eq:delta_r}
\delta \bm r&= \bm r - \bar{\bm r}\\ \label{eq:delta_q}
\delta \bm q &= \bm q \otimes \bar{\bm q}^*.
\end{align}
\end{subequations}
Then, adopting a linearization technique similar
to~\cite{Lefferts-Markley-Shuster-1982, Pittelkau-2001} one can
linearize \eqref{eq:dot_q} about the nominal quaternion $\bar{\bm
q}$ and nominal velocity
\[ \bar{\bm\omega}=\bm\omega_{g} + \bar{\bm b}_{g},\]
to obtain
\begin{equation} \label{eq:diff_delqv}
\diff \delta \bm q_v = - \bar{\bm\omega} \times \delta \bm q_v +
\frac{1}{2} \delta \bm b_{g} + \frac{1}{2} \bm\epsilon_{g}.
\end{equation}
Note that, since $\delta q_o$ is not an independent variable and it
has variations of only the second order, its time derivative can be
ignored, as suggested in~\cite{Lefferts-Markley-Shuster-1982}.

Similarly, the equation of translational motion \eqref{eq:diff_v}
can be linearized about the nominal trajectory obtained from
\begin{equation}\label{eq:diff_barv}
\ddot{\bar{\bm r}} = \bm A(\bar{\bm q}) \bar{\bm a} - \bm g,
\end{equation}
where $\bar{\bm a} = \bm a + \bar{\bm b}_a$. Finally, using
\eqref{eq:A_delq} in differentiation \eqref{eq:delta_r} with respect
to time yields
\begin{align} \notag
\diff \delta \dot{\bm r} &= \bm A(\delta\bm{q} \otimes \bar{\bm{q}}) \big( \bar{\bm a} + \delta \bm b_a + \bm\epsilon_a \big) - \bm A(\bar{\bm q}) \bar{\bm a}  \\
\label{eq:diff_delr}& \approx -2 \bar{\bm A} [\bar{\bm a} \times]
\delta \bm q_v + \bar{\bm A} \delta \bm b_a + \bar{\bm A}
\bm\epsilon_a ,
\end{align}
where $\bar{\bm A} \triangleq \bm A(\bar{\bm q})$.  Note that in
derivation of \eqref{eq:diff_delr}, the second- and higher-order
terms of $\delta \bm q_v$ are ignored. Then, assuming
time-invariant parameters, one can assemble equations
 \eqref{eq:dot_rho}, \eqref{eq:dotb_g}, \eqref{eq:diff_delqv}, and \eqref{eq:diff_delr}
in the standard state space form as
\begin{equation} \label{eq:linsyscont}
\delta \dot{\bm x} = \bm F \delta \bm x + \bm G \bm\epsilon,
\end{equation}
where vector $\bm\epsilon^T= [\bm\epsilon_g^T \; \bm\epsilon_{b_g}^T
\; \bm\epsilon_a^T \; \bm\epsilon_{b_a}^T]\in \mathbb{R}^{12}$
contains the entire process noise; and
\begin{subequations}
\begin{equation}\label{eq:F}
\bm F = \begin{bmatrix}
- [ \bar {\bm\omega} \times] & \bm 0_3 & \bm 0_3 &  \frac{1}{2} \bm 1_3  & \bm 0_3 &  \bm 0_{3\times 3n} \\
\bm 0_3 &  \bm 0_3  & \bm 1_3 &  \bm 0_3 & \bm 0_3 &   \bm 0_{3\times 3n}\\
\bm -2 \bar{\bm A}[\bar{\bm a} \times] & \bm 0_3 & \bm 0_{3\times3}
& \bm 0_3 & \bar{\bm A} &
\bm 0_{3\times3n}\\
& & & \bm 0_{(6+3n)\times(15+3n)}  & &\\
\end{bmatrix}
\end{equation}

\begin{equation} \label{eq:G}
\bm G = \begin{bmatrix} \frac{1}{2} \bm 1_3 & \bm 0_3 & \bm 0_3 &
\bm 0_3
\\ \bm 0_3 & \bm 0_3 & \bm 0_3 & \bm 0_3 \\  \bm 0_3 & \bm 0_3  &  \bar{\bm A}
& \bm 0_3 \\\bm 0_3 & \bm 1_3 & \bm 0_3 & \bm 0_3 \\ \bm 0_3 & \bm 0_3 & \bm 1_3 & \bm 0_3 \\
& \bm 0_{3n\times 12} &
\end{bmatrix}
\end{equation}
\end{subequations}

%------------------------------------------------------
\section{Observability}
\label{sec:Observability}
%------------------------------------------------------
A successful use of Kalman filtering requires that the system be
observable\cite{Aghili-2010p}. A linear time-invariant (LTI) systems is said to be {\em
globally observable} if and only if its observability matrix is full
rank. If a system is observable, the  estimation error becomes only
a function of the system noise, while the effect of the initial
values of the states on the error will asymptotically vanish. The
original observation model, \eqref{eq:obs_nonlin}, and a part of the
process model, \eqref{eq:dot_q}, are nonlinear systems. For
nonlinear system, Hermann {\em et al.} proposed a rank condition
test for ``local weak observability'' of nonlinear system that
involves {\em Lie derivative} algebra \cite{Hermann-Krener-1977}.
Although this technique has been applied for observability of 2D
SLAM \cite{Lee-Wijesoma-Guzman-2006}, the analysis is too complex to
be useful for the 3D case. The observability analysis can be
simplified if the state-space is composed of the errors in terms of
$\delta \bm x$. In that case, the time-varying system \eqref{eq:H}
and \eqref{eq:linsyscont} can be replaced by a piecewise constant
system for observability analysis
\cite{Goshen-Meskin-Bar-Itzhack-1992,Bryson-Sukkarieh-2008}. The
intuitive motion is that such a time-varying system can be
effectively approximated by a pieces-wise contact system without
loosing the characteristic behavior of the original system
\cite{Goshen-Meskin-Bar-Itzhack-1992,Aghili-2010d}.

Now assume that $\bm F_j$ be $\bm H_j$ are the $j$th time segment of
the system's state transition matrix and observation model,
respectively. Then, the observability matrix associated with
linearized system \eqref{eq:linsyscont} together with the
observation model \eqref{eq:H} is
\begin{equation} \notag
\bm{{\cal O}}_j =  \begin{bmatrix} \bm H_j^T & (\bm H_j \bm F_j)^T
& \cdots & (\bm H_j \bm F_j^{3n+14})^T \end{bmatrix}^T .
\end{equation}
The states of the system is instantaneously
observable\footnote{Instantaneous observability means that the
states over time period $[t_{j-1}, \; t_j]$ can be estimated from
the observation data over the same
period\cite{Bryson-Sukkarieh-2008}.} if and only if
\begin{equation} \label{eq:rankO}
\mbox{rank}~\bm{{\cal O}}_j=  3n+15
\end{equation}
which is equivalent to $\bm{{\cal O}}_j$ having  $3n+15$
independent rows. In the following analysis, we remove the index
$_j$ from the corresponding variables for the sake of simplicity
of the notation. Now, we can construct the submatrices of the
observability matrix as

\begin{subequations}\label{eq:HHF}
\begin{align}  \label{eq:HF}
\bm H \bm F &= \begin{bmatrix} -2 [\hat{\bm p}_1 \times]
[\bar{\bm\omega}\times] & \bm 0_3 & -\bar{\bm A}^T & [\hat{\bm p}_1
\times]
& \bm 0_3 & \bm 0_{3 \times 3n} \\
\vdots & \vdots & \vdots  & \vdots &
\vdots & \vdots \\
-2  [\hat{\bm p}_m \times] [\bar{\bm\omega}\times] & \bm 0_3 &
-\bar{\bm A}^T & [\hat{\bm p}_m \times] & \bm 0_3 & \bm 0_{3
\times 3n}
\end{bmatrix} \\ \label{eq:HF2}
\bm H \bm F^2 &=\begin{bmatrix} 2 [\hat{\bm p}_1 \times]
[\bar{\bm\omega}\times]^2 + 2[\bar{\bm a}\times] & \bm 0_3 & \bm 0_3
& -[\hat{\bm p}_1 \times][\bar{\bm\omega}\times] & \bm 1_3 & \bm
0_{3 \times 3n}\\
\vdots & \vdots & \vdots & \vdots & \vdots & \vdots \\
2 [\hat{\bm p}_m \times] [\bar{\bm\omega}\times]^2 + 2[\bar{\bm
a}\times] & \bm 0 & \bm 0 & -[\hat{\bm p}_m
\times][\bar{\bm\omega}\times] & \bm 1_3 & \bm 0_{3 \times 3n}
\end{bmatrix}
\end{align}
\end{subequations}
We consider the observability of the SLAM when there are three
fixed landmarks, i.e., $r=3$. Then, as shown in the
Appendix~\ref{apdx:MRO}, the following {\em block-triangular
matrix} can be constructed from the observability matrix by few
elementary Matrix Row Operations (MRO)
\begin{equation} \label{eq:O_Dleta}
\small \bm{{\cal O}} \stackrel{\text{MRO}}{\longrightarrow}
\bm{{\cal O}}_{\Delta} \!\! = \!\! \begin{bmatrix}
2\bm\Pi &  \bm 0_3  &\bm 0_3 & \bm 0_3 & \bm 0_3 & \bm 0_3 & \cdots & \bm 0_3\\
\times & -\bar{\bm A}^T  &\bm 0_3 & \bm 0_3 & \bm 0_3 & \bm 0_3 & \cdots & \bm 0_3\\
\times & \times &  -\bar{\bm A}^T  &  \bm 0_3 & \bm 0_3 & \bm 0_3 & \cdots & \bm 0_3\\
\times & \times & \times & \bm\Pi & \bm 0_3 & \bm 0_3 & \cdots & \bm 0_3\\
\times & \times & \times & \times &  \bm 1_3 & \bm 0_3 & \cdots & \bm 0_3\\
\times & \times & \times &\times & \times &  \bar{\bm A}^T & \cdots & \bm 0_3\\
\times & \times & \times &\times & \times & \times & \ddots & \bm 0_3\\
\times & \times & \times &\times & \times & \times & \times &
\bar{\bm A}^T
\end{bmatrix}.
\end{equation}
where
\begin{equation} \label{eq:Delta_def}
\bm\Pi = [\bm e_1 \times ]^2 + [\bm e_2 \times ]^2,
\end{equation}
is constructed from the landmark baseline vectors expressed in the
vehicle frame as
\begin{equation}\label{eq:Deltap}
\bm e_i = \frac{\hat{\bm p}_i - \hat{\bm p}_3}{\| \hat{\bm p}_i -
\hat{\bm p}_3\|} \qquad i=1,2.
\end{equation}
The block triangular matrix \eqref{eq:O_Dleta} is full rank if all
of the block diagonal matrices are full rank. Therefore, the
observability condition rests on showing that matrix $\bm\Pi$ is
full-rank.

\begin{proposition} \label{prop:rank_Delta}
If the three fixed landmarks are not place on a straight line, i.e.,
\begin{equation} \label{eq:Deltap_cross}
\bm e_1 \times \bm e_2 \neq \bm 0
\end{equation}
then matrix $\bm\Pi$ is full-rank and so is the observation matrix
$\bm{{\cal O}}$.
\end{proposition}
{\sc Proof:} In a proof by contradiction, we show that $\bm\Pi \in
\mathbb{R}^{3 \times 3}$ must be a full-rank matrix. If $\bm\Pi$ is
not full-rank, then there must exist a non-zero vector $\bm\lambda
\neq \bm 0$ such that
\begin{equation} \label{eq:Deltaeta}
\bm\Pi \bm\lambda = \bm 0
\end{equation}
In view of definition \eqref{eq:Delta_def} and the following
identity,
\begin{equation} \notag
[\bm e_i \times ]^2 = \bm e_i \bm e_i^T - \| \bm e_i \|^2 \bm 1_3,
\end{equation}
one can easily show that \eqref{eq:Deltaeta} is equivalent to
\begin{equation} \label{eq:elambda}
\big( \bm e_1 \bm e_1^T +  \bm e_2 \bm e_2^T \big) \bm\lambda = 2
\bm\lambda.
\end{equation}
Since $\bm e_1 \bm e_1^T$ and $\bm e_2 \bm e_2^T$ are projection
matrices to the subspace $\bm e_1$ and $\bm e_2$, respectively, the
only possibility for the identity \eqref{eq:elambda} to happen is
that $\bm e_1
\parallel \bm e_2 \parallel \bm\lambda$. However, this is a
contradiction because $\bm e_1$ and $\bm e_2$ are not parallel and
therefore matrix $\bm\Pi$ must be full rank.

\subsection{Observability of Piecewise Constant Equivalent}
Now assume that $\bm H_j$ and $\bm F_j$ vary from segment $j=1$ to
$j=r$. Then, the system \eqref{eq:H}-\eqref{eq:linsyscont} is to be
completely observable if the {\em total observability matrix}
\begin{equation}
\tilde{\bm{{\cal O}}} = \begin{bmatrix} \bm{{\cal O}}_1 \\ \bm{{\cal O}}_1 e^{\bm F_1 t_{\Delta_1}}\\\vdots \\
\bm{{\cal O}}_r e^{\bm F_{r-1} t_{\Delta_{r-1}}} \cdots e^{\bm F_1
t_{\Delta_1}}
\end{bmatrix}
\end{equation}
is full rank \cite{Goshen-Meskin-Bar-Itzhack-1992}. Moreover, if
\begin{equation} \label{eq:null_Oj}
\text{null}(\bm{{\cal O}}_j) \subset \text{null}(\bm F_j) \qquad
\forall j=1,\cdots,r
\end{equation}
then it has been shown that $\text{null}(\tilde{\bm{{\cal
O}}})=\text{null}(\tilde{\bm{{\cal O}}}_s)$, where
\begin{equation} \label{eq:som}
\tilde{\bm{{\cal O}}}_s \triangleq [ \bm{{\cal O}}_1^T \; \cdots \;
\bm{{\cal O}}_r^T]^T
\end{equation}
is the {\em stripped observability matrix}
\cite{Goshen-Meskin-Bar-Itzhack-1992}. If the condition
\eqref{eq:Deltap_cross} is satisfied, then the corresponding
observability matrices are full rank, i.e., $\text{null}(\bm{{\cal
O}}_j)=\emptyset$. Consequently, condition \eqref{eq:null_Oj} in
trivially satisfied and the stripped observability matrix
\eqref{eq:som} is full rank. The above development can be summarized
in the following Remark:
\begin{remark}
Assume that linearized system \eqref{eq:H}-\eqref{eq:linsyscont} is
piecewise constant for every single segment $j=1,\cdots,r$. Then,
the system during the time interval $t_1\leq t \leq t_r$ is
completely observable if at least  three known landmarks which are
not placed in a straight line are observed
\end{remark}

%------------------------------------------------------
\section{Adaptive SLAM} \label{sec:Estimator}
%------------------------------------------------------

\subsection{Discrete-Time Model}

The equivalent discrete-time model of \eqref{eq:linsyscont} is
\begin{equation} \label{eq:discrete_model}
\delta \bm x_{k+1} = \bm\Phi_k \delta \bm x_k +  \bm w_k
\end{equation}
where $\bm\Phi_k= \bm\Phi(t_k , t_{\Delta}) $ is the state
transition matrix over time interval $t_{\Delta}=t_{k+1}-t_k$. In
order to find a closed form solution for the state transition
matrix, we need to have the nominal values of some of the states.
Assuming that the nominal values of the bias parameters are obtained
from the latest estimate update, i.e., $\bar{\bm b}_{g_k}= \hat{\bm
b}_{g_k}$ and $\bar{\bm b}_{a_k}= \hat{\bm b}_{a_k}$, the nominal
angular velocity and linear acceleration can be obtained by
averaging the IMU signals at interval $t_k < t \leq t_k +
t_{\Delta}$, i.e.,
\begin{subequations}
\begin{align}
\bar{\bm\omega}_k &= \hat{\bm b}_{g_k} + \frac{1}{t_{\Delta}}
\int_{t_k}^{t_k +  t_{\Delta}} \bm\omega_g( \xi) d \xi \\
\bar{\bm a}_k &= \hat{\bm b}_{a_k} + \frac{1}{t_{\Delta} }
\int_{t_k}^{t_k +  t_{\Delta}} \bm a( \xi) d \xi \qquad 0< \tau \leq
t_{\Delta}
\end{align}
\end{subequations}
Then, the state transition matrix takes on the form:
\begin{equation*}
\bm\Phi(t_k, \tau) =  \mbox{diag} \big(\bm\Phi'(t_k; \tau), \bm
1_{3+3n} \big)
\end{equation*}
where
\begin{equation*}
\bm\Phi'(t_k,\tau) = \begin{bmatrix}
\bm\Lambda_{1_k}(\tau) & \bm 0_3 & \bm 0_3 & \frac{1}{2} \bm\Lambda_{2_k}(\tau)\\
\bm 0_3 & \bm 0_3  &  \bm 1_3 \tau & \bm 0_3 \\
-\bar{\bm A}_k [\bar{\bm a}_k \times] \bm\Lambda_{2_k}(\tau) & \bm
0_3 &  \bm 0_3 & \frac{1}{2} \bm\Lambda_{3_k}(\tau)
\\\bm 0_3 & \bm 0_3 & \bm 0_3 & \bm 1_3
\end{bmatrix}.
\end{equation*}
Here, $\bar{\bm A}_k \triangleq \bm A(\bar{\bm q}_k)$ and the
submatrices of the above matrix are given as
\begin{align} \notag
\bm\Lambda_{1_k}(\tau) =& \bm 1_3 - \frac{\sin \varpi_k
\tau}{\varpi_k}[\bar{\bm\omega}_k \times] + \frac{1 - \cos \varpi_k
\tau}{\varpi_k^2}[\bar{\bm\omega}_k \times]^2
\\ \notag \bm\Lambda_{2_k} (\tau)=&  \bm 1_3 \tau +
\frac{\cos \varpi_k \tau -1}{\varpi_k ^2}[\bar{\bm\omega}_k \times]
+ \frac{\varpi_k \tau  - \sin \varpi_k \tau}{\varpi_k^3}
[\bar{\bm\omega}_k \times]^2
\\ \label{eq:Psi} \bm\Lambda_{3_k}(\tau) =&  \frac{1}{2} \bm 1_3
\tau^2 + \frac{\sin \varpi_k \tau  - \varpi_k \tau }{\varpi_k^3}
[\bar{\bm\omega}_k \times]  +  \frac{ \cos \varpi_k \tau +
 \frac{1}{2} \varpi_k^2 \tau^2
-1 }{\varpi_k^4}[\bar{\bm\omega}_k \times]^2,
\end{align}
where
\begin{equation} \notag
\varpi_k \triangleq \| \bar{\bm\omega}_k \| .
\end{equation}

Presumably, the continuous process noise of the entire system is
with covariance
\begin{equation} \notag
E[\bm\epsilon \; \bm\epsilon^T]=\bm\Sigma_{\epsilon}= \mbox{diag}(
\sigma_g^2 \bm 1_3, \sigma_{b_g}^2 \bm 1_3, \sigma_a^2 \bm 1_3,
\sigma_{b_a}^2 \bm 1_3).
\end{equation}
Then, the covariance matrix of the discrete-time process noise,
which will be used by the KF, can be calculated from
\begin{align} \label{eq:Q_int}
\bm Q_k &= E[\bm w_k \bm w_k^T] = \int_{t_k}^{t_k+t_{\Delta_k}}
\bm\Phi(t)\bm G \bm\Sigma_{\epsilon} \bm G^T \bm\Phi^T(t) {\rm d}t,
\\ \notag &=\mbox{diag} \big( \bm Q'_k , \bm 0_{(6+3n) \times (6+3n)} \big)
\end{align}
where $\bm Q'_k$ is the nonzero sub-matrix of covariance matrix,
which has the following structure
\begin{equation}
\bm Q'_k = \begin{bmatrix} \bm Q'_{k_{11}} & \times & \times & \times & \times  \\
\bm 0_3 & \frac{1}{3} \sigma_a^2t_{\Delta}^3  \bm 1_3  & \times & \times & \times \\
\bm Q'_{k_{31}} &  \bm 0_3  & \bm Q'_{k_{33}} & \times & \times \\
\bm Q'_{k_{41}}  & \bm 0_3 &  \bm Q'_{k_{43}} & \sigma_{b_g}^2
t_{\Delta} \bm 1_3 & \times\\
\bm 0_3 & \bm 0_3 & \bm 0_3 & \bm 0_3 & \sigma_{b_a}^2 t_{\Delta}
\bm 1_3 &
\end{bmatrix}
\end{equation}
Here the symmetric entries of the covariance matrix $\bm Q'_k$ are
not written for the sake of notation simplicity. For small angle
\[\theta_k = \varpi_k \tau \ll 1,  \qquad 0\leq \tau \leq t_{\Delta}\]
 we can say $\sin \theta_k
\approx \theta_k - \frac{1}{6} \theta_k^3 $ and $\cos \theta_k
\approx 1- \frac{1}{2}\theta_k^2$. Consequently, using the Taylor
expansion of the sinusoidal functions in the state transition matrix
\eqref{eq:Psi} and then substituting the reduced order matrix (given
in Appendix~\ref{apdx:Lambdas}) into \eqref{eq:Q_int} and ignoring
the third and higher-order terms of $\theta_k$ after integration
will result in
\begin{align*}
\bm Q'_{k_{11}} & = \Big( \frac{\sigma_{g}^2}{4} t_{\Delta} +
\frac{\sigma_{bg}^2}{12} t_{\Delta}^3 \Big) \bm 1_3 +
\frac{\sigma_{bg}^2}{240}  t_{\Delta}^5
[\bar{\bm\omega}_k \times]^2\\
\bm Q'_{k_{31}} & = \frac{\sigma_{b_g}^2}{32}t_{\Delta}^4 \bm 1_3 -
\frac{\sigma_g^2}{8}t_{\Delta}^2 \bar{\bm A}_k [\bar{\bm a}_k
\times] + \Big(\frac{\sigma_{b_g}^2}{240}t_{\Delta}^5 \bm 1_3 -
\frac{\sigma_g^2}{24}t_{\Delta}^3 \bar{\bm A}_k [\bar{\bm a}_k
\times] \Big)[\bar{\bm\omega}_k \times] \\
&+\Big(\frac{\sigma_{b_g}^2}{576}t_{\Delta}^6 \bm 1_3 -
\frac{\sigma_g^2}{96}t_{\Delta}^4 \bar{\bm A}_k [\bar{\bm a}_k
\times] \Big)[\bar{\bm\omega}_k \times]^2\\
\bm Q'_{k_{33}} & = \frac{\sigma_{b_g}^2}{80} t_{\Delta}^5 \bm 1_3 -
\frac{\sigma_{b_g}^2}{12} t_{\Delta}^3 [\bar{\bm A}_k\bar{\bm a}_k
\times] + \frac{\sigma_{b_g}^2}{2016}t_{\Delta}^7 [\bar{\bm\omega}_k \times]^2 - \frac{\sigma_g^2}{240}t_{\Delta}^5\big( [\bar{\bm A}_k\bar{\bm a}_k \times] [\bar{\bm A}_k\bar{\bm\omega}_k \times]\big)^2\\
\bm Q'_{k_{41}} & =\frac{\sigma_{b_g}^2}{4}t_{\Delta}^2\bm 1_3
+\frac{\sigma_{b_g}^2}{48}t_{\Delta}^4[\bar{\bm\omega}_k \times]^2
\\
\bm Q'_{k_{43}} & =\frac{\sigma_{b_g}^2}{240}t_{\Delta}^3 \Big( 20
\bm 1_3 + 5 t_{\Delta}[\bar{\bm\omega}_k \times] +
t_{\Delta}^2[\bar{\bm\omega}_k \times]^2 \Big).
\end{align*}
In the derivation of the above equations we use the following
identities $\bar{\bm A} \bar{\bm A}^T = \bm 1_3$ and $\bar{\bm A} [
\bar{\bm a} \times] \bar{\bm A}^T = [(\bar{\bm A} \bar{\bm a})
\times]$.

\subsection{Estimator Design}

Before we pay attention to the KF estimator design, it is important
to point out that only the variation of the quaternion, $\delta \bm
q_{v_k}$, and not the quaternion itself, $\bm q_k$, is estimated by
the KF. Nevertheless, the full quaternion can be obtained from the
former variables if the value of the nominal quaternion $\bar{\bm
q}(t_k)$ is given, i.e.,
\begin{equation} \label{eq:deltaqv-}
\delta \hat{\bm q}_k^- = \hat{\bm q}_k^- \otimes \bar{\bm q}^*(t_k)
\end{equation}
For the linearization of the quaternion to make sense, the nominal
quaternion trajectory, $\bar{\bm q}(t)$, should be close to actual
one as much as possible. A natural choice for {\em a posteriori}
nominal value of quaternion at $t_{k-1}$ is its update estimate,
i.e., $\bar{\bm q}(t_{k-1}) = \hat{\bm q}_{k-1}$. Since the nominal
angular velocity $\bar{\bm\omega}_{k}$ is assumed constant at
interval $t_{k-1} \leq t \leq t_{k}$, then according to
\eqref{eq:dot_q} the nominal quaternion evolves from its initial
value $\bar{\bm q}(t_{k-1})$ to its {\em a priori} value $\bar{\bm
q}(t_k)$ by
\begin{equation} \label{eq:qbar_exp}
\bar{\bm q}_k \triangleq \bar{\bm q}(t_k) = e^{\frac{t_{\Delta}}{2}
[\underline{\bar{\bm\omega}}_k \otimes]} \hat{\bm q}_{k-1},
\end{equation}
which will be used at the innovation step of KF. It can be shown
that the above exponential matrix function has the following
closed-form expression
\begin{equation*}
e^{\frac{t_{\Delta}}{2} [\underline{\bar{\bm\omega}}_k \otimes]} =
\big(\cos \frac{\varpi_{k}t_{\Delta}}{2} + \sin \frac{\varpi_{k}
t_{\Delta} }{2} \big) \bm 1_4 + \Big( \frac{2}{\varpi_{k}t_{\Delta}}
\sin\frac{\varpi_{k}t_{\Delta}}{2} - \frac{1}{2} \cos
\frac{\varpi_{k}t_{\Delta}}{2} \Big)
[\underline{\bar{\bm\omega}}_{k} \otimes]
\end{equation*}

The EKF-based observer for the associated noisy discrete system
\eqref{eq:discrete_model} is given in two steps: ($i$) estimate
correction and ($ii$) estimation propagation. The estimate
correction process begins by calculating the filter gain matrix as
\begin{subequations} \label{eq:stepi}
\begin{equation} \label{eq:K_est}
\bm K_k  = \bm P_k^ - \bm H_k^T \big(\bm H_k \bm P_k^- \bm H_k^T+
\bm R \big) ^{-1}
\end{equation}
Next, the states of KF and the covariance matrix are updated in the
innovation step. Recall that only the vector part of the quaternion
variation, not the full quaternion, is included in the KF state
vector. Therefore, we use a priori value of the nominal quaternion
$\bar{\bm q}_k \triangleq \bar{\bm q}(t_k)$ from expression
\eqref{eq:qbar_exp} first to calculate the {\em a priori} quaternion
deviation and then, after the state update in the innovation step,
{\em a posteriori} quaternion deviation is recombined with the
nominal quaternion to obtain the quaternion update. That is
\begin{equation} \label{eq:Innovation}
\begin{bmatrix} \delta \hat{\bm q}_{v_k} \\ \hat{\bm y}_k \end{bmatrix} = \begin{bmatrix}\mbox{vec}(\hat{\bm
q}_k^- \otimes \bar{\bm q}^*_k) \\ \hat{\bm y}_k^-
\end{bmatrix} + \bm K_k \big(\bm z_k - \bm h_k(\hat{\bm x}_k^-)
\big)
\end{equation}

\begin{equation} \label{eq:q_hat}
\hat{\bm q}_k  = \delta \hat{\bm q}_k \otimes
\bar{\bm q}_k=\begin{bmatrix} \delta \hat{\bm q}_{v_k} \\
{\sqrt{1 - \norm{\delta \hat{\bm q}_{v_k}}^2}} \end{bmatrix}
e^{\frac{t_{\Delta}}{2} [\underline{\bar{\bm\omega}}_{k} \otimes]}
\hat{\bm q}_{k-1}
\end{equation}

Note that $\delta \hat{\bm q}_{v_k}^-=\mbox{vec}(\hat{\bm q}_k^-
\otimes \bar{\bm q}^*_k)$ in \eqref{eq:Innovation} is a priori
estimation of the quaternion deviation, where $\mbox{vec}(\cdot)$
returns the vector part of a quaternion. The process of quaternion
update can be summarized as
\[ \hat{\bm q}_k^- \stackrel{\eqref{eq:qbar_exp}}{\longrightarrow} \delta \hat{\bm q}_k^-
\stackrel{\eqref{eq:Innovation}}{\longrightarrow} \delta \hat{\bm
q}_k \stackrel{\eqref{eq:q_hat}}{\longrightarrow} \hat{\bm q}_k
\]
The covariance matrix is updated according to
\begin{equation}
\bm P_k = \big(\bm 1_{15+3n} - \bm K_k \bm H_k \big) \bm P_k^-,
\end{equation}
\end{subequations}

In the second step, the states and the covariance matrix are
propagated into the next time step. Combining \eqref{eq:dot_q} and
\eqref{eq:diff_v}, we then describe the state-space model of the
system as
\begin{equation} \notag
\dot{\bm x} = \bm f(\bm x, \bm\epsilon),
\end{equation}
which can be used for estimating propagation of the states. Thus
\begin{subequations} \label{eq:stepii}
\begin{align}\label{eq:state-prop}
\hat{\bm x}_{k+1}^- &= {\hat{\bm x}_k} + \int_{t_{k}}^{t_{k}+t_{\Delta}} \bm f(\bm x(t), \bm 0)\,{\rm d}t\\
\bm P_{k+1}^- &= \bm\Phi_k \bm P_k \bm\Phi_k^T + \bm Q_k\\ \notag
&=\begin{bmatrix} \bm\Phi_k^{\prime}\begin{bmatrix} \bm P_{11} & \bm
P_{12}^T \end{bmatrix} \bm\Phi_k^{\prime T}+ \bm Q'_k & \times \\
\begin{bmatrix} \bm P_{12} & \bm P_{22} \end{bmatrix}
\bm\Phi_k^{\prime T} & \bm P_{22},
\end{bmatrix},
\end{align}
\end{subequations}
where $\bm P_{11}\in\mathbb{R}^{12\times12}$ and the other
submatrices are obtained from adequate partitioning of the
covariance matrix, i.e.,
\[ \bm P = \begin{bmatrix} \bm P_{11} & \bm P_{12}^T \\
\bm P_{12} & \bm P_{22} \end{bmatrix}. \]

\subsection{Initialization and Landmark Augmentation}

Since  the vehicle frame $\{ {\cal B} \}$ coincides with inertial
frames $\{ {\cal A} \}$ at $t=0$, we can say $\bm r(0)=\bm
0_{3\times 1}$ and $\bm A(0)= \bm 1_3$. Moreover, since $\bm z(0)=
\bm\rho(0) + \bm v$, the initial value of the states can be
adequately chosen as
\[\bm x(0)= \begin{bmatrix}\bm 0_{15 \times 1}  \\ \bm z(0) \end{bmatrix}, \qquad \bar{\bm A}(0) = \bm 1_3, \]
while the initial value of the filter covariance matrix can be
specified as
\[ \bm P^-(0)=\begin{bmatrix} \bm 0 & \bm 0 \\
\bm 0 & \bm R(0) \end{bmatrix} \]

The KF estimation proceeds according to
\eqref{eq:stepi}-\eqref{eq:stepii} cycle as long as the landmarks
are reobserved. However, if a new landmark is observed then the KF
states and its covariance matrix have to be augmented. Let us assume
$\bm z_{\rm new}$ represent the observation associated with a new
landmark at location $\bm\rho_{\rm new}$. Then, the explicit
expression of the new landmark position can be obtained from the
inverse kinematics of the observation as
\begin{align*}
\bm\rho_{\rm new} &= \bm A(\delta \bm q \otimes \bar{\bm q})(\bm z_{\rm new} - \bm v_{\rm new}) + \bm r \\
& \approx \bar{\bm A}(\bm 1_3 + 2 [\delta {\bm q}_v \times])(\bm
z_{\rm new} - \bm v_{\rm new}) + \bm r \\
& \approx \bar{\bm A} \bm z_{\rm new} + \hat{\bm r} \underbrace{ -
\bar{\bm A} \bm v_{\rm new} - 2 \bar{\bm A}[\bm z_{\rm new} \times]
\delta \tilde{\bm q}_v + \tilde{\bm r}}_{\rm noise}
\end{align*}
where $\tilde{\bm q}_v$ and $\tilde{\bm r}$ are the corresponding
estimation errors and $\bm R_{\rm new}=E[\bm v_{\rm new} \bm v_{\rm
new}^T]$. Consequently, the states and covariance matrix are
augmented as
\begin{equation*}
\hat{\bm x}_{\rm new} = \begin{bmatrix} \hat{\bm x} \\ \bar{\bm A}
\bm z_{\rm new} + \hat{\bm
r} \end{bmatrix} \quad \mbox{and} \quad \bm P_{\rm new} = \begin{bmatrix} \bm P & \bm\Upsilon^T \\
\bm\Upsilon & \bar{\bm A}\bm R_{\rm new} \bar{\bm A}^T \end{bmatrix}
\end{equation*}
where
\begin{equation*}
\bm\Upsilon = \begin{bmatrix}-2 \bar{\bm A}[\bm z_{\rm new} \times]
& \bm 1_3 & \bm 0_{3\times(9+3n)} \end{bmatrix} \bm P.
\end{equation*}

\subsection{Noise-Adaptive Filter} \label{sec:noise-adaptive}

Efficient implementation of the KF requires the statistical
characteristics of the measurement noise \cite{Aghili-Kuryllo-Okouneva-English-2010b,Aghili-Parsa-2008b}. The IMU noises can be
either derived from the sensor specification or empirically tuned.
However, the landmark sensor noise is uncertain as they may vary
from one point to the next. Therefore, it is desirable to weight
camera measurement data in the fusion process heavily only when a
``good'' observation data is available. This requires readjusting
the covariance matrix associated with $\bm v$ in the filter's
internal model based upon information obtained in real time from the
measurements.

In a noise-adaptive Kalman filter, the issue is that, in addition to
the states, the covariance matrix  of the measurement noise  has to
be estimated~\cite{Maybeck-1982,Chui-Chen-1998-p113}. Let us define
the residual error
\begin{equation} \notag
\bm\varrho_k \triangleq \bm z_k - \bm H_k \hat{\bm x}_k^-.
\end{equation}
Then the following identity holds
\begin{equation} \notag
\bm\varrho_k = \bm H_k(\hat{\bm x}_k - \hat{\bm x}_k^-) + \bm v_k.
\end{equation}
Taking variance of on both sides of the above equation gives
\begin{equation} \notag \bm R_k = \bm W_k - \bm H_k \bm P_k^- \bm H_k^T \quad
\mbox{with} \quad \bm W_k=E[\bm\varrho_k \bm\varrho_k^T]
\end{equation}
The above equation can be used to estimate the measurement
covariance matrix $\hat{\bm R}_k$ from an ergodic approximation of
the covariance of the zero-mean residual $\bm\varrho$ in the sliding
sampling window with length $w$. That is
\begin{subequations}
\begin{align} \label{eq:S_batch}
\hat{\bm W}_k & \approx
\frac{1}{w}\sum_{i=k-w}^k \bm\varrho_i \bm\varrho_i^T\\
\label{eq:S_recursive}&= \hat{\bm W}_{k-1} + \frac{1}{w} \Big(
\bm\varrho_k \bm\varrho_k^T - \bm\varrho_{k-w}\bm\varrho_{k-w}^T
\Big).
\end{align}
\end{subequations}
where $w$ is chosen empirically to give some statistical smoothing.

%------------------------------------------------------
\section{A Case Study}
\label{sec:simulation}
%------------------------------------------------------

%=============================================================
\begin{figure}[h]
\centering \includegraphics[width=10cm]{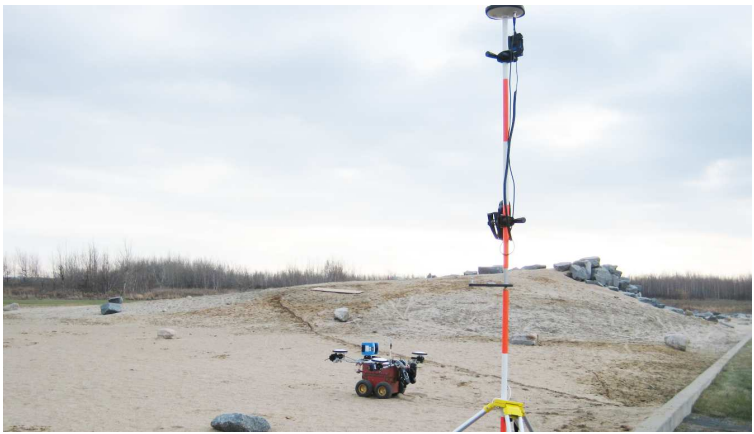} \caption{CSA Mars
emulation terrain.} \label{fig:rover_MET}
\end{figure}
%=============================================================

A series of case studies were conducted on the CSA's red-rover
traversing the $30~\times~60$~m Mars Emulation Terrain (MET), as
shown in Fig.~\ref{fig:rover_MET}, in order to demonstrate the
convergence property of the 3D SLAM with respect to different
numbers of fixed landmarks. The rover is equipped with IMU plus
three RTK GPS antennas, which allows to measure not only the vehicle
position but also its attitude using the method described in
\cite{Aghili-Salerno-2009}. Consequently, the vehicle pose
trajectories obtained from the GPS system are considered as the
``ground-truth'' path.

%---------------------------------------------------------------
\begin{figure}
\centering{\includegraphics[clip,width=9cm]{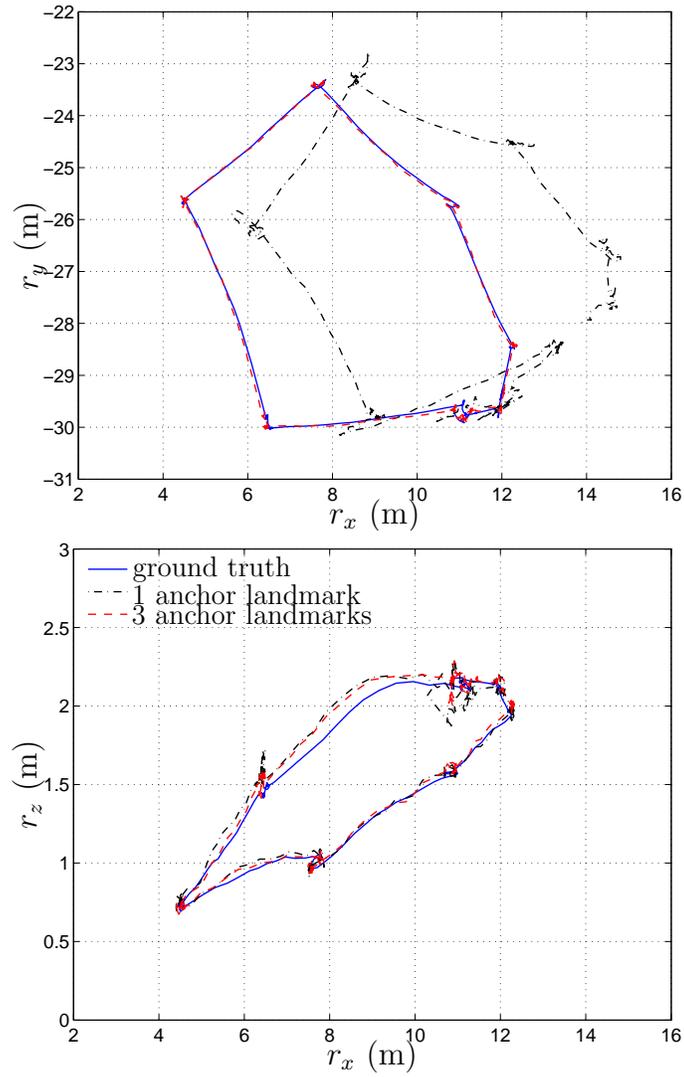}} \caption{3D path
taken by the vehicle.} \label{fig:path}
\end{figure}
%---------------------------------------------------------------

%---------------------------------------------------------------
\begin{figure}
\centering{\includegraphics[clip,width=10cm]{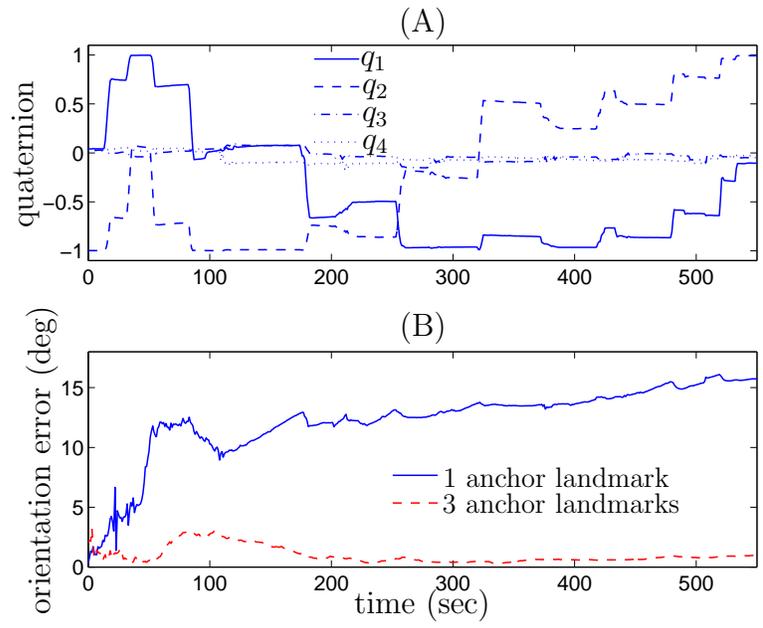}}
\caption{Estimation of vehicle attitude.} \label{fig:quaternions}
\end{figure}
%---------------------------------------------------------------

%---------------------------------------------------------------
\begin{figure}
\centering{\includegraphics[clip,width=10cm]{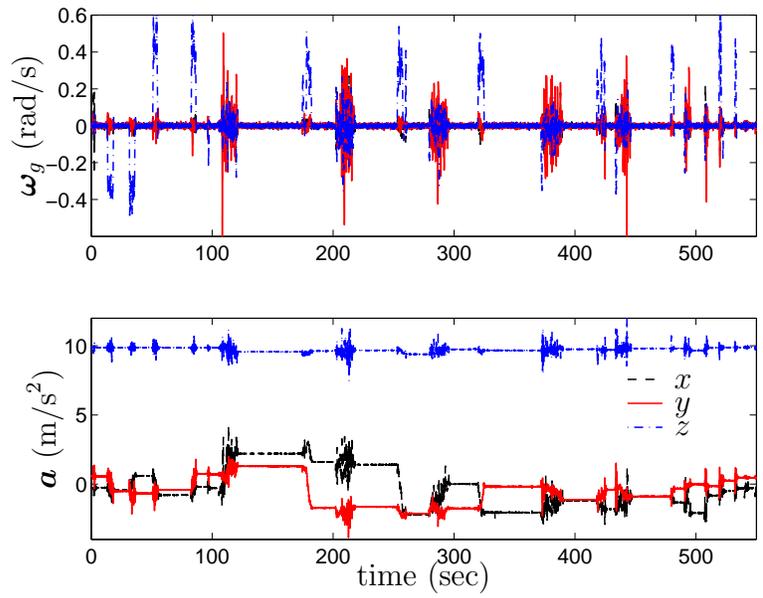}} \caption{IMU
outputs.} \label{fig:imu_out}
\end{figure}
%---------------------------------------------------------------

%---------------------------------------------------------------
\begin{figure}
\centering{\includegraphics[clip,width=10cm]{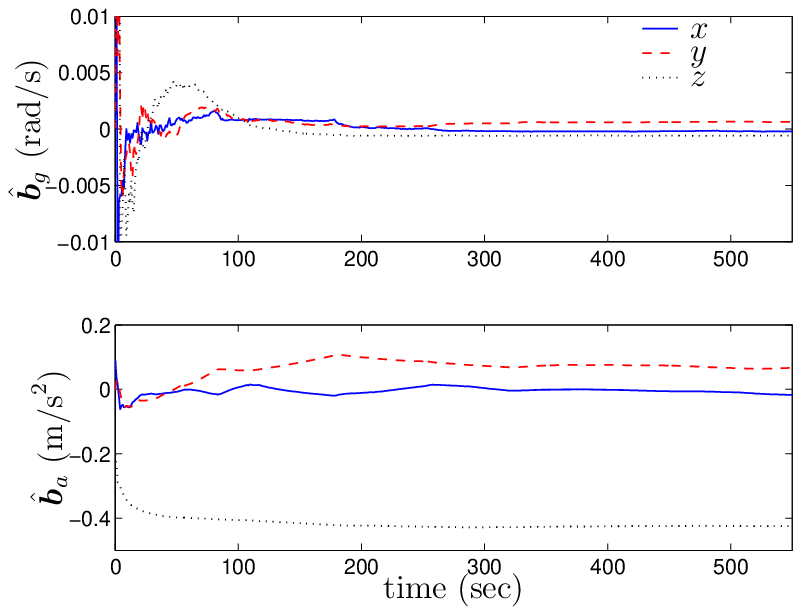}}
\caption{Estimation of the IMU calibration parameters.}
\label{fig:bias}
\end{figure}
%---------------------------------------------------------------

Figs.~\ref{fig:path} and \ref{fig:quaternions}A show the 3D path
taken by the mobile robot and its attitude, respectively, while the
vehicle passing through seven via points. The IMU measurements are
received at the rate of 20~Hz, and the corresponding trajectories
are depicted in Figs.~\ref{fig:imu_out}. In this case study, the
relative positions of the landmarks are simulated by using the
ground truth trajectories and a set of random noise sequences with
difference standard deviations bounded by $0.1 < \sigma_{p_i} <
0.25$~(m). The filter processed the data from three scenarios as:
(i) one of the observed landmarks used as a global reference, (ii)
three of the observed landmarks used as a global reference.
Figs.~\ref{fig:path} shows trajectories of the position estimates
for the two cases, while the corresponding orientation estimate
errors are shown in \ref{fig:quaternions}B. Here, the orientation
errors is calculated by
\begin{equation} \label{eq:attitude_err}
\text{Orientation Error} =  2\sin^{-1}\| \text{vec}( \hat{\bm
q}^*\otimes \bm q_{\rm ref} )\|.
\end{equation}
It is evident from the graphs that the filter did not converge if
only one of the first observed landmark is used as the global
reference. However, the results clearly show that the pose estimates
very well converges to the actual values if three of the observed
landmarks are used as the global references. Fig.~\ref{fig:bias}
illustrates the time history of the estimates of the gyroscope bias
and the accelerometer bias.

%---------------------------------------------
\section{Conclusions}
%---------------------------------------------
Development and the corresponding observability analysis of an adaptive
SLAM by fusing IMU and landmark sensors for autonomous driving vehicles have been
presented. Examining the observability of such SLAM technique in
3-dimensional environment led to the conclusion that the system is
observable if at least three known landmarks which are not placed in
a straight line are observed. The IMU calibration parameters and the
covariance matrix of measurement noise associated with landmark
sensors were estimated upon the sensor information obtained in real
time so that the adaptive estimator is continuously  ``tuned'' as well as
possible.

\appendix

%-------------------------------------------------------------
\section{Reducing $\bm{{\cal O}}$ Through Matrix Row Operations} \label{apdx:MRO}
%-------------------------------------------------------------
The following matrix
\begin{equation} \label{eq:O2}
\small \bm{{\cal O}} \stackrel{\text{MRO}}{\longrightarrow}
\begin{bmatrix}
2[(\hat{\bm p}_1 - \hat{\bm p}_3) \times] &  \bm 0_3  &\bm 0_3 & \bm 0_3 & \bm 0_3 & \bm 0_3 & \cdots & \bm 0_3\\
2[(\hat{\bm p}_2 - \hat{\bm p}_3) \times] &  \bm 0_3  &\bm 0_3 & \bm 0_3 & \bm 0_3 & \bm 0_3 & \cdots & \bm 0_3\\
\times & -\bar{\bm A}^T  &\bm 0_3 & \bm 0_3 & \bm 0_3 & \bm 0_3 & \cdots & \bm 0_3\\
\times & \times &  -\bar{\bm A}^T  & [\hat{\bm p}_1 \times] & \bm 0_3 & \bm 0_3 & \cdots & \bm 0_3\\
\times & \times & \bm 0_3 &[(\hat{\bm p}_1 - \hat{\bm p}_3) \times]   & \bm 0_3 & \bm 0_3 & \cdots & \bm 0_3\\
\times & \times & \bm 0_3 &[(\hat{\bm p}_2 - \hat{\bm p}_3) \times]  & \bm 0_3 & \bm 0_3 & \cdots & \bm 0_3\\
\times & \times & \times & \times &  \bm 1_3 & \bm 0_3 & \cdots & \bm 0_3\\
\times & \times & \times &\times & \times &  \bar{\bm A}^T & \cdots & \bm 0_3\\
\times & \times & \times &\times & \times & \times & \ddots & \bm 0_3\\
\times & \times & \times &\times & \times & \times & \times &
\bar{\bm A}^T
\end{bmatrix}
\end{equation}
can be constructed via the following elementary operations: The
first and second rows of the above matrix are obtained by
subtracting the third row from the first and second rows of matrix
$\bm H$, \eqref{eq:H}. Performing the same operation in matrix $\bm
H \bm F$, in \eqref{eq:HF}, produces the fifth and sixth rows of the
above matrix. The third, forth, seventh rows are picked from first
rows of matrices $\bm H$, $\bm H \bm F$, and $\bm H \bm F^2$,
respectively. The last rows of the above matrix are picked from the
$r+1$th to the $m$th rows of matrix $\bm H$. Now, pre-multiplying
the first and second rows of \eqref{eq:O2} by $[\bm e_1 \times]/\|
\hat{\bm p}_1 - \hat{\bm p}_3 \|$ and $[\bm e_2 \times]/\| \hat{\bm
p}_2 - \hat{\bm p}_3 \|$, respectively, and the adding the resultant
rows and performing similar operations on the fifth and sixth rows
of \eqref{eq:O2} yields
\begin{equation} \label{eq:O3}
\small
\begin{bmatrix}
2\bm\Pi &  \bm 0_3  &\bm 0_3 & \bm 0_3 & \bm 0_3 & \bm 0_3 & \cdots & \bm 0_3\\
\times & -\bar{\bm A}^T  &\bm 0_3 & \bm 0_3 & \bm 0_3 & \bm 0_3 & \cdots & \bm 0_3\\
\times & \times &  -\bar{\bm A}^T  & [\hat{\bm p}_1 \times] & \bm 0_3 & \bm 0_3 & \cdots & \bm 0_3\\
\times & \times & \bm 0_3 & \bm\Pi & \bm 0_3 & \bm 0_3 & \cdots & \bm 0_3\\
\times & \times & \times & \times &  \bm 1_3 & \bm 0_3 & \cdots & \bm 0_3\\
\times & \times & \times &\times & \times &  \bar{\bm A}^T & \cdots & \bm 0_3\\
\times & \times & \times &\times & \times & \times & \ddots & \bm 0_3\\
\times & \times & \times &\times & \times & \times & \times &
\bar{\bm A}^T
\end{bmatrix}.
\end{equation}
Finally, pre-multiplying the forth row of \eqref{eq:O3} by
$-[\hat{\bm p}_1 \times] \bm\Pi^{-1}$ and then add the resultant row
to the third row of \eqref{eq:O3} yields \eqref{eq:O_Dleta}.

%-------------------------------------------------------------
\section{Simplified $\bm\Lambda_i$s'} \label{apdx:Lambdas}
%-------------------------------------------------------------
\begin{align} \label{eq:Psi_simple}
\bm\Lambda_{1_k}(\tau) \approx & \bm 1_3 - \tau [\bar{\bm\omega}_k
\times] + \frac{\tau^2}{2} [\bar{\bm\omega}_k \times]^2 \\ \notag
\bm\Lambda_{2_k}(\tau) \approx & \bm 1_3 \tau - \frac{\tau^2}{2}
[\bar{\bm\omega}_k \times] + \frac{\tau^3}{6} [\bar{\bm\omega}_k
\times]^2\\ \notag \bm\Lambda_{3_k}(\tau) \approx & \bm 1_3
\frac{\tau^2}{2} - \frac{\tau^3}{6} [\bar{\bm\omega}_k \times] +
\frac{\tau^4}{24} [\bar{\bm\omega}_k \times]^2
\end{align}

%-------------------------------------------------------------
\bibliographystyle{IEEEtran}
%\bibliography{references}
%\end{document}

%-------------------------------------------------------------

\end{document}